\pdfoutput=1
\documentclass[11pt]{article}

\usepackage[preprint]{acl}

\usepackage{times}
\usepackage{latexsym}
\usepackage{makecell}
\usepackage{multirow}
\usepackage{hyperref}

\usepackage[T1]{fontenc}

\usepackage[utf8]{inputenc}

\usepackage{microtype}

\usepackage{inconsolata}

\usepackage{graphicx}

\title{Measuring the Effect of Disfluency\\ in Multilingual Knowledge Probing Benchmarks}

 \author{Kirill Semenov \and Rico Sennrich \\
         University of Zurich \\
         \texttt{kirill.semenov@uzh.ch} \\}

\begin{document}
\maketitle
\begin{abstract}

For multilingual factual knowledge assessment of LLMs, benchmarks such as MLAMA use template translations that do not take into account the grammatical and semantic information of the named entities inserted in the sentence. This leads to numerous instances of ungrammaticality or wrong wording of the final prompts, which complicates the interpretation of scores, especially for languages that have a rich morphological inventory. In this work, we sample 4 Slavic languages from the MLAMA dataset and compare the knowledge retrieval scores between the initial (templated) MLAMA dataset and its sentence-level translations made by Google Translate and ChatGPT. We observe a significant increase in knowledge retrieval scores, and provide a qualitative analysis for possible reasons behind it. We also make an additional analysis of 5 more languages from different families and see similar patterns. Therefore, we encourage the community to control the grammaticality of highly multilingual datasets for higher and more interpretable results, which is well approximated by whole sentence translation with neural MT or LLM systems.\footnote{The dataset and all related code is published at the Github repository: \href{https://github.com/ZurichNLP/Fluent-mLAMA}{ZurichNLP/Fluent-mLAMA}.}
\end{abstract}

\section{Introduction}
\label{introduction}

\begin{figure}%
\centering
\includegraphics[width=0.47\textwidth]{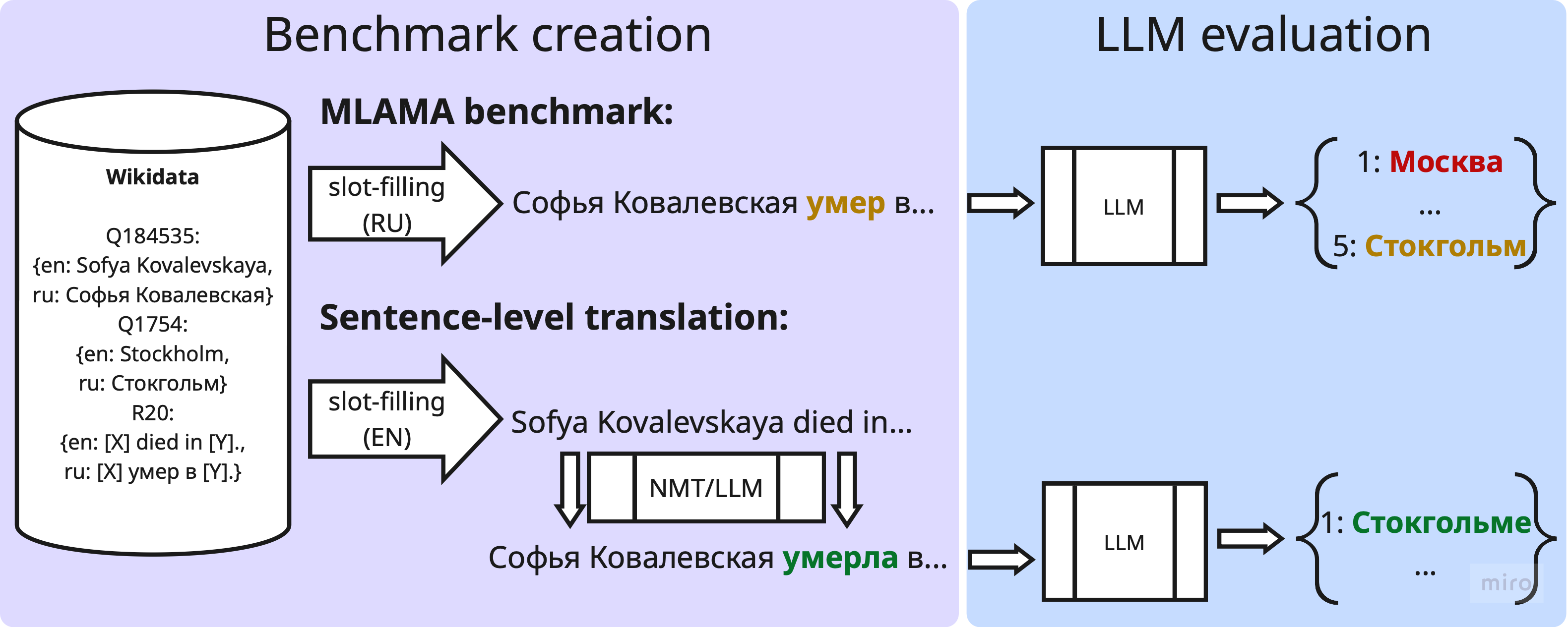}
\caption{Main idea of our experiment. Factual dataset MLAMA uses the templated sentences in each language to prompt LLMs for factual knowledge. This results in many ungrammatical prompts: for Sofya Kovalevskaya, who was the first woman to earn a doctorate in mathematics, the verb "was born" had to be inflected in feminine gender, not in the templated masculine. This leads to poor factual extraction. Sentence-level translation leads to grammatically and lexically coherent sentences and increases factual retrieval of objects (in the correct grammatical forms, in this case, locative case of the word "Stockholm"). The incorrect entities are marked red, the correct but ungrammatical words are in yellow, the correct and grammatical words are in green.}
\label{fig:main}
\end{figure}

Large Transformer-based \citep{vaswani_attention_2017} pre-trained language models (LMs) are known for their fact memorization \citep[see][]{petroni_language_2019,zhong_factual_2021}. For behavioral analysis of factual knowledge retrieval of LLMs, datasets that contain facts from a particular database are used. Several datasets have been created for multilingual LM evaluation \citep[for instance][]{jiang_x-factr_2020,gao_multilingual_2024}. One of the richest datasets (both in terms of language variety and number of factual instances) is MLAMA \citep{kassner_multilingual_2021}, which has been widely used by other researchers and served as a basis for further versions of multilingual knowledge benchmarks such as BMLAMA \citep{qi_cross-lingual_2023}. While being created in the "age" of encoder LMs, it successfully survived the reorientation of the field towards the decoder-only LMs, and is being used as a probe there (see Section~\ref{literature} for examples). 

However, if we take a closer look at the benchmarks and at the formulations of the prompts in languages other than English, we will frequently stumble upon ungrammatical sentences or wrong lexical choice. This stems from the dataset creation pipeline, which included translating the relation types and the named entities separately. Although such a procedure was clearly aimed at uniformity of the prompts, and authors report that manual template correction for German, Hindi and Japanese only had small effects on results, we revisit this question, motivated by numerous fluency issues in the data, and the shift towards decoder-only LLMs which might show different patterns when faced with disfluent prompts. How meaningful are the resulting scores in the face of disfluency? Does a gap in scores compared to English indicate a lack of cross-lingual knowledge transfer, or do disfluent queries act as a confound that explain part of the gap?

To answer this question, we have sampled nine languages and compare factual retrieval accuracy of a Llama-2 model between existing templated sentences from MLAMA and automatic sentence-level translations. As a first set, we consider four Slavic languages which have a considerable range of grammatical categories (such as gender, case and tense markers): Russian, Czech, Ukrainian and Croatian. The languages exhibit diversity in terms of writing systems (Latin for Czech and Croatian, Cyrillics for Russian and Ukrainian), as well as their level of resourcedness. For these languages, consider two translation systems: Google NMT or (in a more controlled manner) ChatGPT with explicit named entity translation. A second set of languages aims to establish generalisation beyond Slavic languages, and includes Spanish, Chinese, Vietnamese, Indonesian and Danish. There, we compare the templated translations of the datasets with the sentence-level Google NMT translations. Our hypothesis for both experiments is that MLAMA (and other template-based benchmarks) substantially underestimate multilingual knowledge retrieval. The brief summarization of our experiment is shown in Figure~\ref{fig:main}.

Our results show significant increase in factual retrieval over most of the nine languages, especially if the model is prompted with Google NMT translation. The increase in performance, however, was not equal for all languages, and the biggest effect (up to 10\%) is observed in Czech, Russian (Slavic), Vietnamese and Indonesian (non-Slavic). 

With this paper, we present the main following contributions:

\begin{enumerate}
    \item We show an empirical evidence of the claim that the grammaticality and lexical accuracy of the prompts matter for the multilingual factual evaluation of the decoder LLMs;
    \item We create and publish the curated version of the MLAMA dataset of the four Slavic and five non-Slavic languages that we used for evaluation;
    \item We publish the straightforward pipeline for extension of the grammatical dataset for the rest of the languages present in MLAMA.%
\end{enumerate}

\section{Related Work}
\label{literature}

Prompt-based probing has been one of the main ways of the factual knowledge analysis of LMs. The first datasets for this task, such as zsRE \citep{levy_zero-shot_2017} or ParaRel \citep{elazar_measuring_2021}, were initially created to test encoder-based models such as BERT \citep{devlin_bert_2018}, but then started being used for decoder models evaluation \citep[for example][]{yao_editing_2023,meng_locating_2022}. Older datasets often serve as a basis for new benchmarks: the LAMA dataset \citep{petroni_language_2019}, another popular factual dataset, was itself an updated version of the T-REx \citep{elsahar_t-rex_2018} dataset. Later, some research teams used LAMA itself as a material for filtered benchmarks like LAMA-UHN \citep{poerner_e-bert_2020}, while others borrowed the principles of data collection for particular domains, such as BioLAMA \citep{sung_can_2021} or LegalLAMA \citep{chalkidis_lexfiles_2023}.

The datasets also demonstrate continuity in terms of multilingual factual knowledge analysis. Just as different multilingual versions of zsRE \citep{wang_retrieval-augmented_2024} and ParaRel \citep{fierro_factual_2022}, the LAMA dataset was translated several times, the most well-known version being MLAMA \citep{kassner_multilingual_2021}. This was a significant contribution to multilingual evaluation of LMs, since the dataset was translated into 53 languages. The MLAMA benchmark has also become the source, whether in terms of material or of creation protocols, to later multilingual datasets, such as BMLAMA \citep{qi_cross-lingual_2023}, GeoMLAMA \citep{yin_geomlama_2022} or DLAMA \citep{keleg_dlama_2023}; it also widely used by the community as is until now \citep{zhao_tracing_2024,xu_language_2023,chen_journey_2023}.

The challenges of prompt-based probing have been a long-standing research focus. Firstly, we cannot be sure that the prompts manually formulated by the researchers (which was the initial approach) would maximize the retrieval of a fact. Several solutions were suggested, from corpus-based retrieval of optimal prompt phrasing \citep{jiang_how_2020}, to gradient-based methods, either by choosing arbitrary tokens for encoders \citep[such as AutoPrompt,][]{shin_autoprompt_2020} or by creating coherent sentences for decoders \citep{yin_benchmarking_2024}. Secondly, since many approaches are based on providing an LLM with a range slot-filling (or next token) options, and ranking the correct options against the distractors, a body of work is related to finding the optimal distractor set: see \citep{khodja_factual_2025} for an overview.

Another set of problems is related to multilingual prompting. DLAMA and GeoMLAMA, but also xGeo and xPeo \citep{gao_multilingual_2024} are datasets that are more balanced in terms of geographical or cultural distribution of facts in the datasets. 
The translation protocol is another axis of variation among the multilingual datasets: the most popular way is using MT models such as Google Translate (see Bi-zsRE \citep{wang_cross-lingual_2024}, FEVER \citep{beniwal_cross-lingual_2024}, MzsRE \citep{wang_knowledge_2023}, sometimes even different NMT model comparison as in mParaRel \citealt{fierro_factual_2022}), followed by using LLMs (as in MULTIGEN \citep{huang_112_2024}, DLAMA, BMIKE-53 \citep{nie_bmike-53_2024} or MLaKE by \citealt{wei_mlake_2024}) and, in rare cases, by explicit rule-based inflection generation with the help of Unimorph-inflect \citep{anastasopoulos_pushing_2019} (such as in X-FACTR \citealt{jiang_x-factr_2020}). Some projects explicitly state that the whole dataset was manually checked after the translation phase (such as GeoMLAMA), others, including MLAMA, state that they covered a subset of languages and relations at the manual check round. Importantly, the procedure of translating datasets with NMT may differ in setups: while datasets like MzsRE translated the full sentences with Google Translate, the MLAMA creators translated only the relation templates (e.g. "[X] was born in [Y]" and then used the translations of the X and Y entities retrieved from WikiData directly, which ignores the grammaticality of the final sentences. 

It is trivial to assume that translation LMs would increase the number of grammatical sentences in the dataset compared to template filling: we expect that based on the training data of both NMT systems and LLMs (which are usually fluent sentences). Consequently, we can see a gradual trend in switching from templates to sentence-level translation by NMT and then by LLMs, while the datasets like MLAMA are still being used in multilingual LLM evaluation. However, to our knowledge, there has been no direct comparison of templated (i.e. not necessarily grammatical or lexically accurate) to sentence-level translated versions of the datasets, hence the question of whether there is a real gain in using fluent sentences for multilingual LLM prompting, or templated sentences are already a sufficient baseline, still stands. With this paper, we attempt to fill this lacuna. 

\section{Method}
\label{method}

We follow the general framing of factual knowledge that can be seen in \citep{elazar_measuring_2021} or \citep{meng_locating_2022}: a \textbf{fact} \(f_i\) is a triplet \(\{s_i, r_i, o_i\}\), where subject \(s_i\) and \textbf{object} \(o_i\) are named entities (NE), and \(r_i\) is a type of relation between the two NEs. The structured fact then has to be formulated as a natural language sentence, which we call \textbf{verbalization} (following \citealt{khodja_factual_2025}). We therefore say that an LLM "knows" the fact \(f_i\) if, prompted with a verbalization of a correct fact \(\{s_i, r_i, o_i\}\) and of the incorrect triplets \(\{s_i, r_i, o_*\}\) differing by object NE, the model assigns higher probability to the correct fact verbalization rather than to the incorrect triplets, which are called "\textbf{distractors}" (also following \citealt{khodja_factual_2025}). In the following subsections, we will cover the fact, verbalization and distractor choice in more detail, as well as final evaluation. 

\subsection{Language, Fact and Distractor Choice} %
\label{method:choice}

The initial MLAMA dataset comprises facts from 53 languages, which cover a broad typological, genealogical and geographical sample. At the same time, the authors note that the templates (not the final verbalizations) were checked manually only for three languages. In this work, since we focus on grammaticality and fluency, we narrow down the scope of languages and chose four Slavic languages, Russian (ru), Czech (cs), Ukrainian (uk) and Croatian (hr), and five non-Slavic languages, Spanish (es), simplified Chinese (zh), Vietnamese (vi), Indonesian (id) and Danish (da), in order to assess the fluency more attentively. A short information about the language families, scripts and resourcedness of the languages is provided in Appendix~\ref{sec:appendix_F}.

The pool of facts in MLAMA comprised 41 relation types mostly from Wikidata \citep{vrandecic_wikidata_2014}. For the analysis, we choose a subset of 15 relation types from Wikidata only, which were unambiguous and had a distractor pool large enough.  For an overview of the included and excluded relations, see Appendix~\ref{sec:appendix_B}\footnote{It is necessary to note, that this limitation is not specific to our work: MLAMA and other highly multilingual factual datasets are not designed for direct cross-lingual comparisons. Very few datasets (like DLAMA by \citealp{keleg_dlama_2023}) aim at balancing the datasets by the number and variety of facts in each language, but such a dataset is significantly harder to create for more than 3-4 languages.

}. 

The pool of subject and object translations is retrieved from Wikidata directly. To enrich the distractor pool (and for the additional experiment in Section~\ref{discussion:aliases}), we extract the aliases for all objects in question from Wikidata (i.e., for the entity denoting New York, we also retrieve its aliases like "NYC", "New York, NY", "The big Apple" etc.).  

The distractors are chosen following the initial MLAMA Typed Querying setting: they need to be of the same entity type as the correct object. For the sake of compute load, we put the maximum boundaries on the number of distractors by sampling 50 entities from the total pool of objects for a particular relation.%

\subsection{Dataset Creation}
\label{method:dataset}
We compare three types of verbalization of the same set of knowledge facts to see if sentence-level translation helps factual retrieval:

\begin{enumerate}
    \item \textbf{Template} filling: the translated template formulations are borrowed from the MLAMA dataset as is and filled with NEs from Wikidata. This is the expected usage scenario for the MLAMA benchmark.
    \item \textbf{GT}: The templates in English are first filled with English names for NEs and then translated with Google Translate. 
    \item \textbf{ChatGPT}: For the experiment with the Slavic languages, we provide the third verbalization: filled English templates (same as in GT) and the translations of the subject and object NEs (retrieved from Wikidata), and prompt GPT-4o-mini\footnote{https://openai.com/index/gpt-4o-mini-advancing-cost-efficient-intelligence/} to translate the full sentence using the provided NE translations. For each relation, we create few-shot instruction (20 examples) to show the model how we want it to use the Wikidata NE translations and how we expect it to use appropriate inflection of the relation verbalization. A few-shot prompt example is shown in Appendix~\ref{sec:appendix_A}. 
\end{enumerate}

We intend to probe an LLM with a part of the verbalization that precedes the object NE. Thus, we split the generated verbalizations into the prompts and the expected endings (object NEs). The template verbalizations are split straightforwardly by cutting the "[Y]." suffix from the end of the sentence; the GT and ChatGPT verbalizations are split with the help of Stanza package \citep{qi_stanza_2020} that searches for the (possibly inflected) object NEs in the sentences. Thus, for each fact, we get up to three different prompts and up to three forms of object NEs (sometimes the generated verbalizations or their parts may not differ); we consider all forms (both non-inflected and possibly inflected) of object NEs as correct ones. While the ChatGPT translation is expected to be consistent in word order of the prompts of the same kind, we cannot control that for Google Translate. Thus, we filter such facts from the dataset for which the GT verbalization differs significantly in word order (e.g. an object NE in the beginning of the sentence). 

\subsection{Experiment Setup}
\label{method:setup}

For each fact \(f_i\), we concatenate the verbalization prompt \(v_i\) and either the set of correct object NE %
or 50 distractors sampled from the same language-relation subset of the dataset (in total, they form the candidate set). Next, we feed the model with such a set of prompts, calculate the log probabilities of tokens corresponding to object NEs, and rank the possible endings of the verbalization by these log probabilities. Then, for a given fact, its score is 1 if the object is ranked among the top $n$ results, and 0 otherwise. We follow the principle of evaluation introduced in \citep{petroni_language_2019}, where it was called Precision at $n$; however, in our case the pool of adversarials is of a fixed size, thus, it is more appropriate to call it recall at $n$, $R_i@n$. To see the stability of the trends in verbalization difference, we generalize $n$ up to 5 in some of our experiments, contrary to MLAMA score that only considers $R@1$.

We calculate the percentage of such hits for a given verbalization type (template, GT or ChatGPT) within a given language-relation subset and note it as \(R@n\). We are interested in comparing the \(R@n\) values within the same language between the verbalization types; thus, for the general results we will average the scores for the whole language dataset. We do not compare the scores between the languages since the sets of facts in different languages differ significantly. This limitation is not specific to our work: MLAMA and other highly multilingual factual datasets are not designed for direct cross-lingual comparisons. Very few datasets (such as DLAMA) aim to balance the datasets by the number and variety of facts in each language, but such a dataset is significantly harder to create for more than 3-4 languages.

We choose the LLaMA-2-7b model \citep{touvron_llama_2023} for our experiment, since this model is one of the few models that explicitly claims its functionality in all four Slavic languages in question. The experiments were run on single NVIDIA GeForce RTX 4090 GPU, the total time of the experiment was 23 hours. 

\section{Results}
\label{results}

\begin{table*}[h]
\centering

\begin{tabular}{|l|rrrr|rrrr|} 
 \hline & \multicolumn{4}{c|}{R@1 $\uparrow$} & \multicolumn{4}{c|}{Mean Rank $\downarrow$} \\
\hline
 Verbalization   &    ru &    cs &    uk &    hr &   ru &   cs &   uk &   hr \\
\hline
 Template        & 0.512 & 0.579 & 0.488 & 0.707 & 4.91 & 4.17 & 5.12 & 2.94 \\
 GT              & 0.586 & 0.67  & 0.525 & 0.704 & 3.95 & 3.16 & 4.43 & 2.68 \\
 ChatGPT         & 0.545 & 0.615 & 0.492 & 0.653 & 4.65 & 3.53 & 5    & 3.02 \\
\hline
\end{tabular}

\caption{Main experiment results: comparison of the performance of three verbalization prompt types. The left columns represent $R@1$ metric (percentage of cases where the LLM ranked the correct object as the most probable; higher score is better retrieval). The right columns show mean value the most probable correct object rank for each fact (lower score is higher rank, therefore better ranking).}
\label{tab:general}
\end{table*}

\subsection{Slavic Languages}

Our first experiment shows (see the left half of Table ~\ref{tab:general}) that the verbalizations generated by Google Translate and by ChatGPT, in general, extract more facts from the LLM than the templated ones. Notably, the GT verbalizations also work considerably better than the ChatGPT-generated ones. In case of the higher-resourced languages (Czech and Russian), the increase in performance is up to 10\% of the retrieved facts (under GT verbalization). The only exception is the language with the lowest resources, Croatian, where the template and GT scores oscillate around the same value with a decrease of ChatGPT performance. We also hypothesized if such an increase in performance is observed only in the $R@1$ setup, or the trend is steady across different top $n$ values. In Appendix~\ref{sec:appendix_C} we demonstrate that GT, followed by ChatGPT, has a steady delta across $n$ up to 5; moreover, in the case of Croatian, the GT-generated verbalizations start extracting slightly more knowledge than the templated ones.

\begin{table}[h]
\centering

\begin{tabular}{|l|rrrr|}
\hline
 Verbalization   &   ru &   cs &   uk &   hr \\
\hline
 Template        & 5.80 & 8.49 & 7.86 & 8.31 \\
 GT              & 8.59 & 7.55 & 9.02 & 9.02 \\
 ChatGPT         & 7.54 & 7.28 & 9.07 & 8.54 \\
\hline
\end{tabular}

\caption{Average rank delta that shows how many ranks the inflected form of ground truth object is higher (closer to rank 1) than its non-inflected counterpart (higher number means bigger delta in favor of the inflected object).}
\label{tab:infl_delta}
\end{table}

Another way of looking at the increase in factual retrieval depending on verbalization is to look at the distributions of the highest correct ranks for each fact. A demonstrative metric would be the mean value of a distribution which is given in the right half of Table ~\ref{tab:general}: we see that, for the high-resourced languages, the GT verbalizations give an increase of the correct answers by one rank, and up to half of the rank for the low-resourced ones. Since the single mean value may not be an informative description of the whole distribution, the more detailed visualizations of the rank distributions are provided in Appendix~\ref{sec:appendix_C}, showing similar outcomes.

The results above were focusing more on different verbalizations of the prompt beginnings. However, we should also remember that the MLAMA framework only counted the non-inflected forms of objects as the correct answers, while we mined the inflected object NE forms from the GT and ChatGPT verbalizations and allowed them as correct together with the non-inflected ones, and kept track of the ranks of both of them. We compare the rank differences between the inflected and non-inflected object forms by selecting 8 relation types where the object is expected to be inflected, and only such occurrences where exactly two forms of the ground truth object were found (assuming that the ground truth form of the templated sentence is the non-inflected form, and another one is the inflected one). Then, we subtract the rank of the non-inflected form from the inflected one and average such differences over language and verbalization type. The results in Table~\ref{tab:infl_delta} show that, firstly, inflected forms are preferred over non-inflected ones in all setups (including the templates), and secondly, compared to templates, the delta in GT verbalization increases from 0.7 ranks (in Croatian) up to almost 3 ranks (in Russian). We therefore see that both the fluency of the beginning of the prompt and the correct inflection of the object NE contributes to increasing the retrievability of a fact from the model.

\subsection{Non-Slavic Languages}

In the experiment above, we restrict ourselves to four languages to manually create few-shot prompts for each relation and then check the resulting translations. Our results show that the biggest increase in retrieval is made with Google Translate verbalization, which does not require human editing. Thus, we decided to run comparison of two verbalizations, template and GT, on five more languages which belong to different language families and use various writing systems (still, we restricted ourselves to only such languages where we can judge the grammaticality of the final outputs post factum). Our sample included Spanish, simplified Chinese, Danish, Vietnamese and Indonesian. Notably, we also sampled a smaller number of relations (namely, eight) than for the first experiment: we selected only the relations that have object in the end of the sentence for all 5 languages (see motivation for that in Section~\ref{discussion:questions}).

\begin{table*}[t]
\centering

\begin{tabular}{|l|rrrrr|rrrrr|} 
 \hline & \multicolumn{5}{c|}{R@1 $\uparrow$} & \multicolumn{5}{c|}{Mean Rank $\downarrow$} \\
\hline
 Verbalization   &    es &    zh &    vi &    id &    da &   es &    zh &   vi &   id &   da \\
\hline
 Template        & 0.725 & 0.02  & 0.587 & 0.547 & 0.568 & 2.64 & 19.98 & 4.1  & 3.7  & 4.16 \\
 GT              & 0.725 & 0.059 & 0.765 & 0.786 & 0.64  & 2.61 & 15.68 & 2.59 & 2.12 & 3.84 \\
\hline
\end{tabular}

\caption{$R@1$ and average rank scores for five additional languages. The notation is the same as in Table~\ref{tab:general}.}
\label{tab:add_langs}
\end{table*}

Table~\ref{tab:add_langs} shows the results of this experiment\footnote{The experiments were run on single NVIDIA GeForce RTX 4090, the total time was 9 hours.}. The results show similar trends as for the Slavic languages: GT verbalization works as an upper bound for all languages, yielding up to 24\% in $R@1$ score and decreasing down to 5 ranks in average rank score. The only exception is Spanish, which remains stable in terms of both the $R@1$ and average rank score (for particular relation types, however, we see increase in $R@1$ up to 5\%). The reasons for such an increase differ depending on language: for Danish and some relations in Spanish, it is proper choice of inflectional category, prepositions or articles; for languages with a narrow inflectional inventory like Vietnamese and Indonesian, the main effect is mostly explained due to more appropriate translations of the relations. With Chinese\footnote{The overall quality of factual retrieval for Chinese is notably low. Our manual check of the results did not show any technical reason for such a poor performance. We hypothesise that this may be related to the poor tokenization quality of the texts written in Chinese characters (compared to Latin and Cyrillic scripts); for general discussion about the multilingual tokenization inequality and its effect on language performance refer to \citep{ahia_all_2023} (which, however, does not include Chinese). We leave the analysis of that hypothesis for the future research.}, most of gains in factual retrieval are explained by either preservation of the Latin spelling of the proper name, or by providing both the Chinese transliteration of the name and the initial Latin spelling in parentheses; we can consider it as one of the forms of explicitation described in Section~\ref{discussion:qualitative}. Overall, we consider this a good sign showing that the sentence-level translation of the dataset can curate a range of language-specific problems which are unrelated to each other.

\section{Discussion}
\label{discussion}

\subsection{Qualitative Analysis}
\label{discussion:qualitative}

One of the most surprising observations in the metrics above was the significant prevalence of GT verbalizations over the ones generated by ChatGPT, despite even 20-shot prompting. The detailed analysis reveals the possible reason behind that: while ChatGPT was forced to follow the subject and object name translations from Wikidata labels and enforce the uniformity of relation formulations, Google Translate was not only free to choose the most common phrasing of the relation, but also to add clarifications for the subject entity type. Such additional information may help LLM to resolve the ambiguities (for instance, when an entity is a movie name which can also be used as a regular phrase, like "The Pianist") and prune it to the correct answer. This behavior of Google Translate can be called "explicitation" and is a well-known technique within translation practice \citep{murtisari_explicitation_2016}. The type of explicitation we see in this case is called "pragmatic explicitation" \citep{klaudy_explication_1998} and is used to help the target audience of a different culture to better understand the text.
Various types of pragmatic explicitation are seen in our dataset, starting with explicit definitions of the entity types, ending with punctuation such as brackets, which goes in line with \citep{adel_metadiscourse_2006} that includes typographical markers into explicitation forms. Notably, even such minor additions as brackets can increase the rank of the retrieved entity to the top $n$ choices.

Another notable observation was comparison of two opposite relations: "[X] is the capital of [Y]" and "The capital of [X] is [Y]" and its verbalizations, in Czech. In both cases, the template translation of the relation was completely wrong due to ambiguity of the term "capital" in English: it was translated as "kapit{\'a}l" (as "capital goods"), not as correct "hlavn{\'i} m{\v e}sto" (as "capital city"). Interestingly, the $R@1$ value for the relation "The capital of [X] is [Y]" (i.e. when the country entity was provided and the model had to guess the capital entity) increased from 0.52 (template verbalization) to 0.89 (GT verbalization), while the opposite relation (where the model had to guess the country given the capital) increased from 0.79 (template) to 0.83 (GT). The disparity in factual retrieval of the opposite relations is known as the "reversal curse" in the LLM evaluation \citep{berglund_reversal_2023} and is observed frequently, and if we were to compare only the template verbalization of the two relations, we would get such an asymmetry; however, we see that the correct translation of the relation "curates" such a disparity and decreases the gap almost to zero. This example also highlights the necessity of sentence-level translation of prompts, since this minimizes the risk of the wrong choice of relation disambiguation in the target language.

\subsection{Named Entity Aliases: Subjects and Objects}
\label{discussion:aliases}

As noted in Section~\ref{method:choice}, both subject and object NEs have different variants of phrasing in Wikidata. For each entity, there is a default way of phrasing it and the alternatives called "aliases". The MLAMA benchmark did not take into account this plurality of NE naming; in other benchmarks, we only know of X-FACTR \citep{jiang_x-factr_2020} counting any of the object aliases as the correct answers, while the subject aliases, to our knowledge, remain not analyzed. The default Wikidata phrasing of the entities is language- and culture-dependent and sometimes can be arguable. For example, the default phrasings of the personal names in the Russian Wikidata are provided in the form of "Surname, First Name (Patronymic if applicable)": "Tolstoi, Lev Nikolaevich", which is formally grammatical but has a strong "bureaucratese" connotation, whereas a formal but more neutral way of addressing will be "First name (Patronymic) Surname": "Lev Nikolaevich Tolstoi". As a more general case, we can think of the translation VS preservation of the movie or book titles: in the Czech and Croatian versions of Wikidata, the translated version is slightly preferred as the default phrasing. 

Our experiment is designed in such a way that in ChatGPT verbalization, the translation of the subject and object NEs was controlled (by providing the default translations from Wikidata into target languages), while in GT setting the system chose the most probable translation of the English NE itself. Thus, we see multiple examples %
of the choice of alternative namings of both subject and object NEs by Google Translate over all relations and all four languages. For example, for the Russian personal name translation, GT chooses more neutral way of addressing.

The analysis of variation of different subject phrasing in GT versus two other verbalization does not show significant effect. The neutral phrasing of the Russian personal names does not yield much for retrieval (compared to formal phrasing in templates and ChatGPT, the increase is from 0.01 to 0.06 score depending on relation type). The preservation of the English book or movie name by GT (instead of translation suggested by Wikidata) also does not show consistent change, with anecdotal evidence of increasing the retrieval quality for Czech and Croatian and decreasing it for Ukrainian. In most cases such a variation goes in hand with the explicitation artifacts described above, which seems to have a bigger effect on fact retrieval. 

Evaluating the variation of all possible subject phrasings would have been too computationally-consuming (as we would have to create prompts for all possible aliases of the prompt for the correct inflection of the relation). With the object NE aliases, we make an experiment by including all possible aliases into the pool of the correct answers (we do not inflect the objects for the same reason of preserving compute by not generating the new sentences; on the other hand, we include the English NE labels to include the probability of activating the representations of the most resourced language). We re-run the evaluation for all languages and relations given this wider pool of correct answers; the results on $R@1$ score can be found in Appendix~\ref{sec:appendix_E}. We see that while all scores increase, the trend remains the same, with GT being the best verbalization, followed by ChatGPT and templates.

Our comparison shows that, despite variation in object and subject NE, the more important factors in factual retrieval are the overall fluency of the prompt and, in some cases, various explicitation techniques. 

\subsection{Language Resourcedness}
\label{discussion:resourcedness}

One of the motivating ideas of our experiments was the expectation that the increase in factual retrieval will be more noticeable for the lower-resourced languages, while the high-resourced languages could manage retrieval well even under distorted input. In our sample, the most resourced language is Russian among Slavic languages, and Spanish among non-Slavic languages (0.13\% of LLaMA-2 training data each). Ukrainian and Vietnamese are seen less (0.07\% and 0.08\% training data, respectively), while the rest of the languages (Czech, Croatian, Indonesian, Danish) are seen in from 0.03\% to 0.01\% training data.

In reality, we did not observe any clear trend: the biggest increase in factual retrieval was for Russian and Czech; the improvement for Ukrainian was smaller, while Croatian showed comparable performance for templated and GT verbalizations. If we look at the sample of languages from our second experiment, the trends become even less clear: Chinese and Spanish are equal to Russian in Llama-2 training data (0.13\% each), and show completely different patterns in factual retrieval. Similarly, Danish and Croatian are almost equally low-resourced in the training data (0.02\% and 0.01\%, respectively), yet Danish benefits from sentence-level translation considerably more.

\subsection{Open Questions}
\label{discussion:questions}
\textbf{1. How to measure grammaticality?} In our experiments, we assume beforehand and then check manually the grammaticality of three types of verbalizations. However, this approach is too time-consuming and is hardly scalable. To approximate the grammaticality of sentences, we computed the log probabilities of different verbalizations for the same fact; this metric was not consistent in terms of showing the fluent sentences. Neither did help surface-based metrics such as chrf \citep{popovic_chrf_2015} that we applied to compare template sentences against GT or ChatGPT verbalizations. Additionally, we applied the grammar error correction (GEC) systems for Russian and Czech such as \citep{rothe_simple_2021} to see whether it would detect grammatical inconsistencies in templated sentences. We observed that the GEC systems mostly targeted spelling mistakes within the words rather than grammatical agreement, therefore such systems also cannot help us with the grammaticality evaluation. Our observation goes in line with the overviews of the GEC systems for languages other than English (such as in \citep{volodina_multiged-2023_2023}.

\begin{figure}[h]
  \centering
  \includegraphics[width=\linewidth]{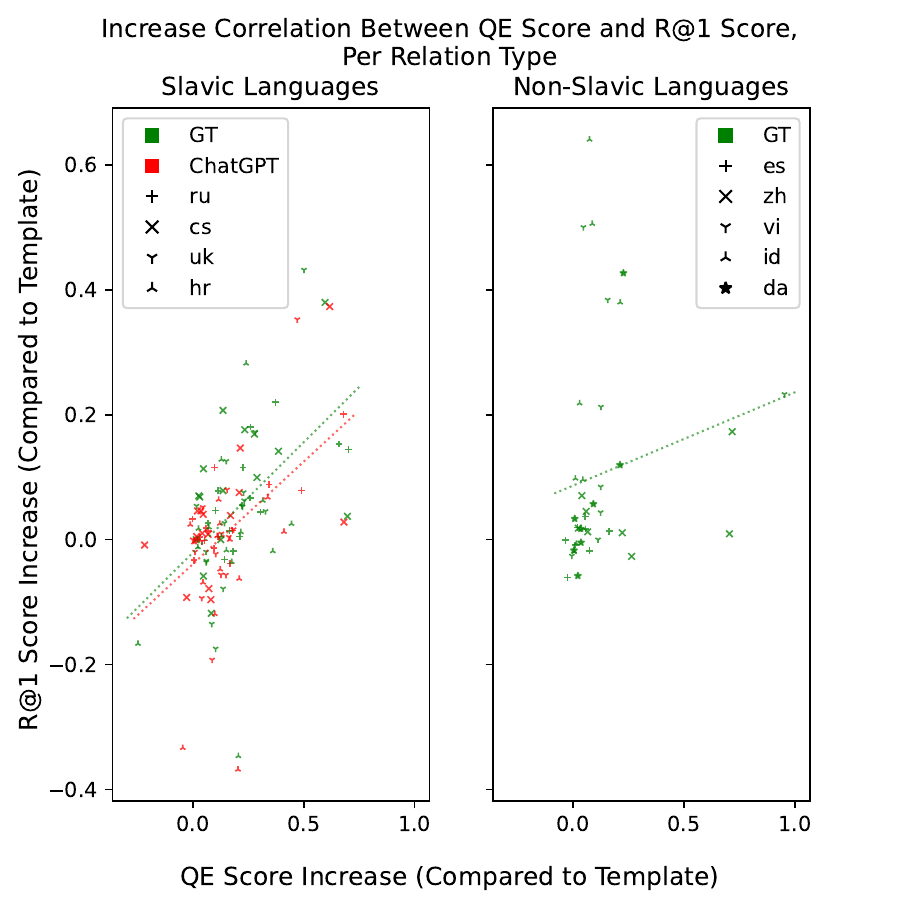}
  \caption{Correlation between increase in QE scores (x-axis) and increase in $R@1$ performance (y-axis) of the GT and ChatGPT verbalizations compared to templates. The left subplot shows statistics for the first experiment (Slavic languages), the right - for the second experiment (non-Slavic languages). Each point represents the averaged value over a particular language, relation and verbalization type. The color represents the verbalization type; the shape of the point stands for the language. The dotted lines show linear regressions fit by all data points at each subgraph for a corresponding type of verbalization.}
  \label{fig:qe_plot}
\end{figure} 

We can also hypothesise that a quality estimation score can be an indirect estimation (QE) of the grammaticality of the sentence (although, according to classical framing of the MT task, it estimates accuracy and fluency of the translation at the same time). Indeed, the scores that we obtained with the COMET model\footnote{\url{https://huggingface.co/Unbabel/wmt20-comet-qe-da}} showed a considerable increase of QE scores (up to 9\% for Slavic and up to 23\% for non-Slavic languages), if we compare the templated translations to GT or ChatGPT outputs. Moreover, the increase of QE scores shows positive correlation with the increase of our target metrics, for instance, $R@1$ (up to 0.8 Pearson's score). This is illustrated in Figure~\ref{fig:qe_plot}, and the detailed information about the observed correlations can be found in Appendix~\ref{sec:appendix_D}. 

Thus, to make grammaticality evaluation scalable, we need to operationalize the notion of grammaticality and create the respective metric for it. %

\textbf{2. How to prompt models multilingually?} We follow the spirit of the MLAMA probe in our experiment design: MLAMA benchmark was created for encoder models, whose main training objectives and inference functions was masked token prediction. Thus, the templated sentences were fed to mBERT with the object token substituted by "[MASK]" sequences. Analogously, since the pre-training objective and inference function for decoder-only models is next token prediction, we formulate the prompts in such a way that the prompt verbalization would contain the full name of the subject NE and the formulation of the relation, while the comparison will be done for the log probabilities of the tokens making up object NEs which are placed in the end of the sentences (e.g., a verbalization would look like "Leo Tolstoi was born in" + {object entities}). However, the languages differ by the so-called basic word order (the most unmarked placing of subject, verb and object in declarative sentences), which is formalized as SVO, VSO etc. (for more information, see the WALS project by \citealt{matthew_dryer_world_2024}). Noticeably, our prompting strategy will work for languages which are O-final; however, there is a large number of languages that are V-final (for example, Turkish, Hindi and Japanese), where the explication of the relation (mostly formulated through a verb phrase) is put after the object. If we apply the existing MLAMA benchmark to such languages for a decoder model in a straightforward way (by splitting the sentence at the place of object), this will not provide the model with information about the relation. Alternatively, we can come up with another prompt schema: showing the whole sentence with a masked object and then ask the model to name the object (e.g. "Leo Tolstoi was born in [Y]. Y:" This may work well in the instruction-tuned versions of the models (especially after few-shot examples), but with this setup we will not be able to disentangle the parametric factual knowledge from the alighment skills learnt through stages like SFT and RLHF. Therefore, we incline towards prompting the model in the way we did (finish the sentence with missing object), but extension of that framework to verb-final languages from the MLAMA pool would require additional curation of the verbalizations so that the objects would be put in the end of the sentence. In most cases, this is possible even for languages with SOV and OSV word orders, as such languages do allow some structures with an object in the end (usually some types of highlighting of one of the constituents), but for this task, careful few-shot prompting through ChatGPT and post-checking by fluent speakers will be required.

\section{Conclusions}
\label{conclusions}

This work analyzes different types of prompting for multilingual factual knowledge evaluation of an LLM depending on whether the prompts are formulated in a fluent manner (translated by an NMT or by an LLM as a whole sentence) or ignoring the grammatical and lexical consistency (filling of the fixed templates). We compare different verbalization for four Slavic and five (typologically and graphically diverse) non-Slavic languages, and show, to our knowledge, the first evidence for an intuitive assumption that sentence-level translations help retrieving more facts from the model than templated translations.
This is an indication that cross-lingual factual retrieval may be underestimated in experiments based on MLAMA or derivative datasets that are template-based. With these results, we encourage the community to use properly translated datasets for higher and more interpretable results of multilingual factual knowledge prompting.

\section*{Limitations}
\label{limitations}

Our work has several limitations:

\textbf{1. Small samples over different dimensions. }The analysis was done for four languages from the same group within the Indo-European family, plus five typologically diverse languages. This narrow focus served to manually create/verify parts of the datasets and to do qualitative analysis.  The relation sample represents less than a half of initial relations from MLAMA. This happened for several reasons: some relations were too ambiguous, too broad in terms of entity types included, had too few object entities, or their template verbalization did not have an object in the end of the prompt. We also tested that on a single LLM since this was the only popular model that explicitly states that it was trained on all languages from our sample. 

\textbf{2. Lack of control in GT.} We could not control the exact content of the sentences that were generated by the NMT system (Google Translate), which resulted in artifacts such as explicitation. This also seems to be the reason for the highest performance of GT compared to a more controlled ChatGPT-generated translations. Yet, even the latter shows increase in the factual retrieval, which is shown both by $R@1$ metric and by ranking distribution of the correct answers. On top of that, due to variation in word order and translation choice of the entities in GT, we had to filter out the facts that were inconsistent with a format of a prompt, which decreased the number of data points in the resulting dataset. 

\textbf{3. Distractor choice.} We sampled the distractors in a pseudo-random reproducible manner (with the help of hashing). However, there is an open discussion on what should be optimal distractor choice \citep{khodja_factual_2025}. Still, we think that our distractor samples (50) are big enough to ensure variety within them, especially since the metric $R@n$ is considering only up to 10\% of the first ranks of the candidate set. 

\textbf{4. Fact choice.} The problem inherited from MLAMA is the imbalance of facts between languages in the initial dataset (up to 10 times between the highest- and the lowest-resourced languages). Therefore, in our final factual sample, Croatian was the smallest with less than 2000 facts, compared to Russian, Czech and Ukrainian with approximately 6600, 3900 and 4100 facts, respectively.  Additionally, the set of entities and relations is West- and specifically Anglo-centric, which was addressed in the datasets like \citep{keleg_dlama_2023}. Still, we believe that our comparison that was done on several thousand facts for each language allows us to highlight the importance of the proper translation of the sentences even for the imbalanced benchmarks.

\section*{Acknowledgments}

This research was funded by NCCR Evolving Language, Swiss National Science Foundation Agreement 51NF40\_180888. We also thank the following people for helping with the preparation of the few-shot prompts and qualitative analysis of the larger sample of languages: Yulia Alpaieva (Charles University) for Ukrainian, Michelle Wastl (UZH) for Croatian, Nam Luu (University of the Basque Country) for Vietnamese, Sophia Conrad (UZH) for Danish, Polina Nalsedskova (HSE University) for Indonesian.

\bibliography{mlama}

\appendix
\onecolumn

\section{Few-Shot Prompt Example}
\label{sec:appendix_A}
Below you can find an example of the few-shot prompt used to generate the ChatGPT verbalization of a fact. A different prompt was created for each language and each relation type. For the sake of brevity, we show here 5 shots of a prompt for Czech language and relation P19 (birthplace of a person); the rest of the 20 prompts in total were formatted the same way. At inference, the model is provided with "Source sentence", "Subject translation" and "Object translation" fields, and is expected to generate the translated sentence only after the word "Translation: " (see the last paragraph on the page). 

Note that the fields "Subject translation" and "Object translation" are extracted from Wikidata and then are forced to be used as translations of the corresponding entities; yet they can be inflected if necessary: n this particular relation, the object entities are put in locative case, e.g. "Lucembursko" (Nominative) -> "Lucembursku" (Locative). Apart from that, the verb is also inflected differently ("narodila se" or "narodil se" depending on gender of the subject). 
\begin{verbatim}
You are a professional English-Czech translator. You are given English sentences 
about subjects and objects. You are also given translations of subjects and objects 
separately. You need to translate full sentences to Czech. When translating, you 
have to use the translated subjects and objects. Pay special attention 
to grammatical agreement between the words in the translated sentences.
When translating, follow the examples:
        
        Source sentence: Cunigunde of Luxembourg was born in Luxembourg .
        Subject translation: Kunhuta Lucemburská
        Object translation: Lucembursko
        Translation: Kunhuta Lucemburská se narodila v Lucembursku.
        
        Source sentence: Karel Schwarzenberg was born in Prague .
        Subject translation: Karel Schwarzenberg
        Object translation: Praha
        Translation: Karel Schwarzenberg se narodil v Praze.

        Source sentence: William Carlos Williams was born in Rutherford .
        Subject translation: William Carlos Williams
        Object translation: Rutherford (New Jersey, U. S.)
        Translation: William Carlos Williams se narodil v Rutherfordu (New Jersey, U. S.).

        Source sentence: Marion Davies was born in Brooklyn .
        Subject translation: Marion Daviesová
        Object translation: Brooklyn
        Translation: Marion Daviesová se narodila v Brooklynu.

        Source sentence: Peter Mark Roget was born in London .
        Subject translation: Peter Roget
        Object translation: Londýn
        Translation: Peter Roget se narodil v Londýně.

        ...
        
        Source sentence: Theodore the Studite was born in Constantinople .
        Subject translation: Theodoros Studijský
        Object translation: Konstantinopol
        Translation: 

\end{verbatim}

\section{Relations Overview}
\label{sec:appendix_B}

\begin{table}[h]
\footnotesize
\setlength{\tabcolsep}{4pt}

\begin{tabular}{|l|l|l|l|l|l|l|l|l|l|l|}
\hline
 \makecell{Relation\\ID}   & en Template                           & ru       & cs       & uk       & hr       & es   & zh       & vi       & id      & da       \\
\hline
 P101                      & [X] works in the field of [Y] .       & 605      & 396      & 396      & 226      & --   & --       & --       & --      & --       \\
 P103                      & The native language of [X] is [Y] .   & 736      & 458      & 445      & 231      & 882  & 554      & $269_{36}$ & 334     & 737      \\
 P108                      & [X] works for [Y] .                   & 91       & 51       & 52       & $18_{28}$  & 149  & $162_{46}$ & $46_{38}$  & $52_{34}$ & 92       \\
 P127                      & [X] is owned by [Y] .                 & 402      & 246      & 258      & 84       & --   & --       & --       & --      & --       \\
 P1376                     & [X] is the capital of [Y] .           & 217      & 214      & 216      & 174      & --   & --       & --       & --      & --       \\
 P159                      & The headquarter of [X] is in [Y] .    & 450      & 240      & 303      & 84       & 483  & 465      & 93       & 221     & 259      \\
 P19                       & [X] was born in [Y] .                 & 493      & 248      & 193      & 103      & 919  & 272      & 86       & 194     & 597      \\
 P20                       & [X] died in [Y] .                     & 679      & 418      & 342      & 158      & --   & --       & --       & --      & --       \\
 P36                       & The capital of [X] is [Y] .           & 614      & 453      & 617      & 294      & 647  & 538      & 361      & 428     & 439      \\
 P364                      & The original language of [X] is [Y] . & 373      & $181_{43}$ & 212      & $64_{33}$  & 394  & 319      & $71_{24}$  & 183     & $228_{44}$ \\
 P407                      & [X] was written in [Y] .              & 583      & 355      & 404      & $163_{42}$ & 702  & 223      & 193      & 287     & 342      \\
 P449                      & [X] was originally aired on [Y] .     & $242_{28}$ & $99_{27}$  & $96_{18}$  & $39_{18}$  & --   & --       & --       & --      & --       \\
 P463                      & [X] is a member of [Y] .              & $151_{45}$ & $54_{40}$  & $130_{30}$ & $102_{29}$ & --   & --       & --       & --      & --       \\
 P495                      & [X] was created in [Y] .              & 414      & 205      & 240      & 81       & --   & --       & --       & --      & --       \\
 P740                      & [X] was founded in [Y] .              & 416      & 256      & 214      & 88       & 782  & 275      & 78       & 160     & 233      \\
\hline
\end{tabular}

\caption{Overview of relations used in the main experiment.}
\label{tab:relation_table}
\normalsize	
\end{table}

Table~\ref{tab:relation_table} shows statistics about the relation types and the sizes of datasets that make up these relations used in the experiments. The "Relation ID" column represents the Wikidata ID, "en Template" column shows the English template that was extracted from the English version of MLAMA (it was then used as a ground for GT and ChatGPT sentence-level translations into the target languages). The columns with language IDs show the number of facts in each dataset that were used in the experiment (i.e. after translation and filtering out of the GT verbalizations which had a wrong word order). By default, the number of distractors that were used for each relation type is equal to 50. In case of difference, this number (which is equal to the whole pool of other entities for a given relation) is provided in subscript for the corresponding relation type and language. 

For the first experiment with the Slavic languages, fifteen relations out of the whole pool were used. These fifteen relations comprise only one third of the relations from the initial MLAMA dataset. The reasons for filtering out other relations were the following:

\begin{enumerate}
    \item The prompt formulation in at least one of the target languages did not have object in the end of the sentence, for example, relation P106 is formulated as "[X] is a[Y] by profession." and similarly in other templates. The reason for excluding such phrasings is provided in Section~\ref{discussion:questions}.
    \item The relation has too few unique object NEs. For example, relation P413 "[X] plays in [Y] position." has fewer than 10 unique objects, which would make the distractor pool too narrow and automatically increase the metrics.
    \item Relations that are too broad and/or ambiguous. For instance, relation P527 "[X] consists of [Y]" includes chemical elements ("water consists of hydrogen"), deities ("Dashavatara consists of Rama"), political entities ("G20 consists of Italy") and so on. Moreover, if another correct object will be in the distractor pool (for instance, "Germany" for the prompt "G20 consists of "), it would not be scored. 
    \item Relations that are poorly formulated in English. Examples include P1001 "[X] is a legal term in [Y]." (which rather means the objects that apply to a particular jurisdiction, such as a parlament or a national flag of a particular county), or P140 "[X] is affiliated with the [Y] religion." (which either means the beliefs of a particular person, or the religion to which a particular temple (church/mosque/gurudwara/etc.) is dedicated). 
    
\end{enumerate}

For the second experiment, due to variation in the word order in typologically differences, we used only eight relations from the pool of the first experiment, such that the objects in the templates are sentence-final (see reason 1 above).

\section{Detailed Statistics of the Experiment With Slavic Languages}
\label{sec:appendix_C}

\begin{figure}[h]
  \centering
  \includegraphics[width=\linewidth]{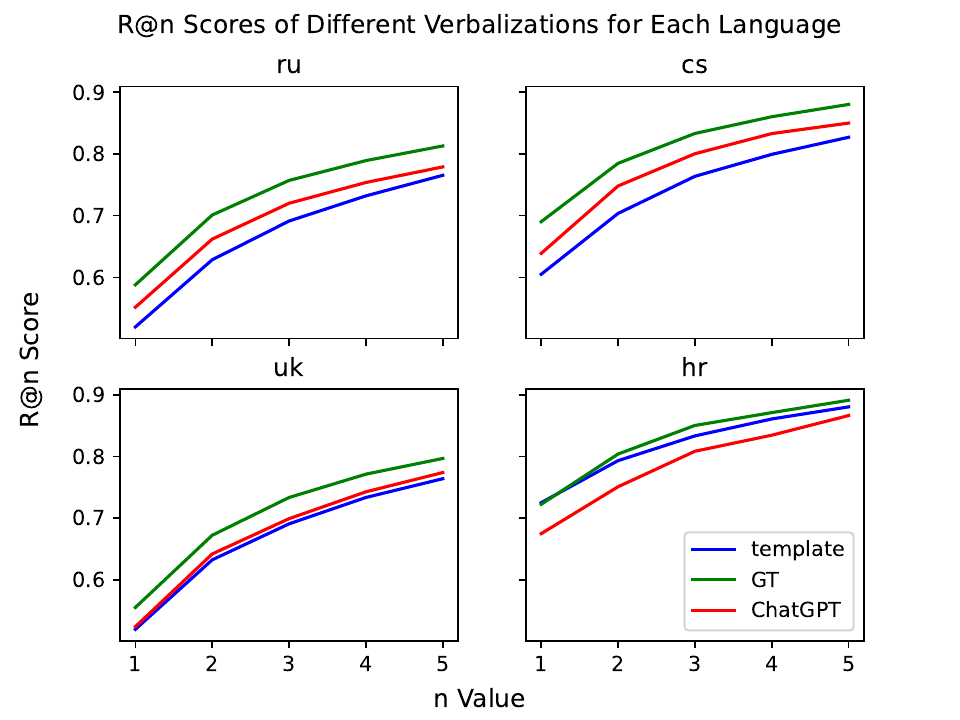}
  \caption{Comparison of $R@n$ scores depending on $n$. Higher score shows better retrieval. }
  \label{fig:general}
\end{figure}

In Figure~\ref{fig:general}, we show the trends in the $R@n$ metric scores for each language depending on the value of $n$. We see that the scores for each verbalization increase consistently, showing the same trend of GT > ChatGPT > template verbalization for each language except Croatian. Thus, we see that the superiority of the sentence-level translation verbalizations is robust against the choice of $n$.

\begin{figure}[h]
  \centering
  \includegraphics[width=\linewidth]{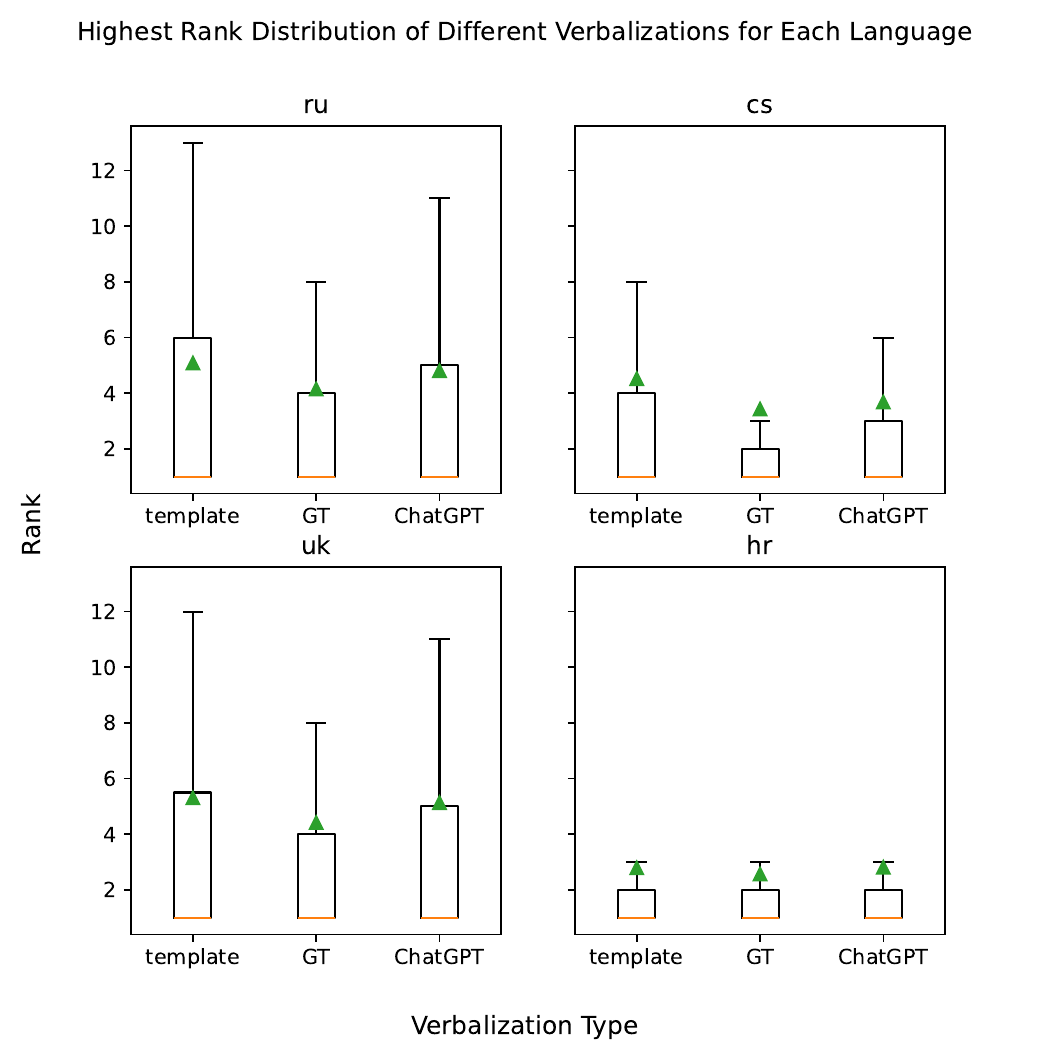}
  \caption{Distribution of the highest correct ranks for each verbalization in each language; lower score is better. For clarity, the box plots do not show outliers, which explain the skew of the average ranks (shown as green triangles) higher than the third quartiles.}
  \label{fig:rank_boxplot}
\end{figure}

In Figure~\ref{fig:rank_boxplot}, we show the box plots that represent the distributions of the highest ranks of the correct object for each verbalization in each language. This gives us a broader view at the distributions compared to a single average rank metric. We see that, apart from being closer to one on average, the ranks of the sentence-level translated verbalizations (especially GT) have a narrower spread compared to templated ones, which makes their performance more stable.

For both graphs (as well as any other score in the paper), we show scores from a single run since we ensured the reproducibility of the ranking by making reproducible sets of correct objects and distractors (by sampling the entities by their hash IDs). 

\section{Grammaticality Estimation}
\label{sec:appendix_D}

\begin{table*}[t]
\setlength{\tabcolsep}{5pt}
\centering

\begin{tabular}{|cl|llll|lllll|}
\hline
 \makecell{Met-\\ric}      & \makecell{Verbali-\\zation}   & ru     & cs     & uk     &     hr & es    & zh    & vi    & id    & da     \\
\hline
 \multirow{2}{*}{$\Delta$} & GT                            & 0.073  & 0.091  & 0.038  & -0.003 & -0.0  & 0.039 & 0.179 & 0.239 & 0.072  \\
                           & ChatGPT                       & 0.032  & 0.036  & 0.005  & -0.054 & N/A   & N/A   & N/A   & N/A   & N/A    \\
\hline
 \multirow{2}{*}{$r$}      & GT                            & 0.608* & 0.498  & 0.788* &  0.26  & 0.436 & 0.408 & 0.135 & 0.592 & 0.838* \\
                           & ChatGPT                       & 0.78*  & 0.613* & 0.741* &  0.206 & N/A   & N/A   & N/A   & N/A   & N/A    \\
\hline
\end{tabular}

\caption{Detailed statistics on QE score difference and its correlation with $R@1$ metric increase, averaged for every language. The QE score difference is calculated by subtracting the QE score of the template verbalization from either GT or ChatGPT verbalization. These scores are shown in the two upper rows of the table (under the $\Delta$ label). In an analogous manner (by subtracting the templated score from the scores from GT or ChatGPT verbalization) we compute the increase in the $R@1$ metric, and calculate the resulting Pearson's correlation between the increase of QE and increase of $R@1$ (to compute the correlation within each language, we firstly average the differences for each relation type). The correlation coefficients are represented in the two lower rows in the table (under the $r$ label); the significant correlation ($p<0.05$) is marked with asterisk. }
\label{tab:qe}
\end{table*}

The only consistent predictor for grammaticality estimation that we could find were the QE metrics. We demonstrate it with the increases in GT and ChatGPT performance (compared to templated translation) both in QE score and in our target $R@1$ metric. In the Table~\ref{tab:qe}, we show the scores of the increases in QE and their correlations to $R@1$, averaged by language. The correlation is also demonstratively shown on a more granular level of the language-relation pairs in Figure~\ref{fig:qe_plot}. 

We also made additional evaluation of grammaticality of the prompts in a more controlled manner by concentrating on specific relation types. Namely, we analyzed the quality of handling the grammatical gender in Slavic languages. For that, we selected two relations, "[X] was born in [Y]" and "[X] died in [Y]", (P19 and P20, respectively). The verbs in the past tense are conjugated by gender (masculine, feminine or neutral). The templated verbalizations always used the masculine form of verb, irrespective of the gender of the person [X]. Therefore we concentrated on the increase in grammatical coherence of the verbalizations verbalizations for the persons of the female gender. For that, we extracted information about their sex/gender from Wikidata, and selected the subset of facts correlating only to women (selecting values "female" or "transgender female"). Then, we counted the percentage of feminine verb forms for each verbalization in each language, as well as the R@1 score for these (only female-related) subsets of facts. The results of this analysis are shown in Table~\ref{tab:gender}. We can see that, firstly, the GT and ChatGPT verbalizations became predominantly accurate in terms of the verb form usage; secondly, for most of the cases, these verbalizations increased the quality of the factual extraction of the corresponding facts. Interestingly, for the subset of facts that are related to women, for most languages, the bigger increase in grammaticality (and in factual retrieval) is connected to ChatGPT verbalizations, not GT ones. 

\begin{table*}[t]
\centering

\begin{tabular}{|l|ll|ll|ll|ll|}
\hline
 \multirow{2}{*}{Verbalization} & \multicolumn{2}{c|}{ru} & \multicolumn{2}{c|}{cs} & \multicolumn{2}{c|}{uk} & \multicolumn{2}{c|}{hr}  \\
                                & \%(F)                   & R@1(F)                  & \%(F)                   & R@1(F)                  & \%(F) & R@1(F) & \%(F) & R@1(F) \\
 Template                       & 0.0                     & 0.373                   & 0.0                     & 0.375                   & 0.0   & 0.311  & 0.0   & 0.545  \\
 GT                             & 80.7                    & 0.38                    & 89.8                    & 0.591                   & 90.2  & 0.246  & 84.8  & 0.515  \\
 ChatGPT                        & 93.3                    & 0.447                   & 97.7                    & 0.466                   & 96.7  & 0.361  & 97.0  & 0.485  \\
\hline
\end{tabular}

\caption{Statistics on subset of relations P19 and P20 ("[X] was born in [Y]" and "[X] died in [Y]"), where X is a woman (and therefore the verb has to be in feminine gender). The statistics are provided for each language separately. The "\%F" column demonstrates the percentage of the verbalizations with female subject where the feminine gender was used correctly; the "R@1(F)" column shows the R@1 score for the subset of the facts with the female subjects.}
\label{tab:gender}
\end{table*}

\section{Additional Experiments}
\label{sec:appendix_E}

\begin{table*}[t]
\centering

\begin{tabular}{|l|rrrr|rrrr|} 
 \hline & \multicolumn{4}{c|}{R@1 $\uparrow$} & \multicolumn{4}{c|}{Mean Rank $\downarrow$} \\
\hline
 Verbalization   &    ru &    cs &    uk &    hr &   ru &   cs &   uk &   hr \\
\hline
 Template        & 0.57  & 0.63  & 0.539 & 0.736 & 3.79 & 3.18 & 4.04 & 2.43 \\
 GT              & 0.633 & 0.709 & 0.583 & 0.736 & 3.27 & 2.49 & 3.52 & 2.29 \\
 ChatGPT         & 0.59  & 0.67  & 0.54  & 0.693 & 3.76 & 2.84 & 4.01 & 2.54 \\
\hline
\end{tabular}

\caption{$R@1$ and average rank scores when aliases are allowed as correct answers. The trends remain the same as in the main results.}
\label{tab:wide}
\end{table*}

Table~\ref{tab:wide} shows the main statistics for the experiment where we allowed Wikidata aliases of objects as correct answers\footnote{The experiments were run on single NVIDIA GeForce RTX 4090, the total time was 25 hours.}. They show increase in both $R@1$ score and average rank; however, the ranking of three types of verbalizations and the delta between them remains almost unchanged. This is an evidence of an importance of the whole sentence translation that cannot be approximated by mere expansion of the correct object phrasing pool.

\section{Language Overview}
\label{sec:appendix_F}

The languages chosen for the experiments (sorted by the mass in training data) are listed below:

First experiment, Slavic languages (subgroups of the Slavic group are provided below): 

\begin{enumerate}
    \item \textbf{Russian}, code: "ru", East Slavic subgroup, Cyrillic script, 0.13\% of pre-training data;
    \item \textbf{Ukrainian}, code: "uk", East Slavic subgroup, Cyrillic script, 0.07\% of pre-training data;
    \item \textbf{Czech}, code: "cs", West Slavic subgroup, Latin script, 0.03\% of pre-training data;
    \item \textbf{Croatian}, code: "hr", South Slavic subgroup, Latin script, 0.01\% of pre-training data. 
\end{enumerate}

Second experiment, non-Slavic languages: 

\begin{enumerate}
    \item \textbf{Spanish}, code: "es", Indo-European (Romance) family, Latin script, 0.13\% of pre-training data;
    \item \textbf{Chinese}, code: "zh", Sino-Tibetan family, Chinese (simplified) script, 0.13\% of pre-training data;
    \item \textbf{Vietnamese}, code: "vi", Austroasiatic family, Latin script, 0.08\% of pre-training data;
    \item \textbf{Indonesian}, code: "id", Austronesian family, Latin script, 0.03\% of pre-training data;
    \item \textbf{Danish}, code: "da", Indo-European (Germanic) family, Latin script, 0.02\% of pre-training data;
\end{enumerate}

\end{document}